\documentclass[a4paper,fleqn]{cas-dc}

\usepackage{hhline}
\usepackage{arydshln}

\usepackage[numbers]{natbib}
\emergencystretch=\maxdimen
\def\tsc#1{\csdef{#1}{\textsc{\lowercase{#1}}\xspace}}
\tsc{WGM}
\tsc{QE}
\tsc{EP}
\tsc{PMS}
\tsc{BEC}
\tsc{DE}
\usepackage{caption}
\usepackage{subcaption}
\begin{document}
\def\floatpagepagefraction{1}
\def\textpagefraction{.001}
\makeatletter
\def\ps@pprintTitle{%
   \let\@oddhead\@empty
   \let\@evenhead\@empty
   \let\@oddfoot\@empty
   \let\@evenfoot\@oddfoot
}
\makeatother

\shorttitle{Pose is all you need: The pose only group activity recognition system (POGARS)}
\shortauthors{Haritha Thilakarathne, Aiden Nibali, Zhen He, Stuart Morgan}

\title [mode = title]{Pose is all you need: The pose only group activity recognition system (POGARS)}
\author[1]{Haritha Thilakarathne}[orcid=0000-0003-1407-8231]
\ead{p.thilakarathne@latrobe.edu.au}
\author[1]{Aiden Nibali}
\author[1]{Zhen He}
\author[2]{Stuart Morgan}
\address[1]{Department of Computer Science and Information Technology, La Trobe University, Melbourne, VIC, Australia}
\address[2]{Australian Institute of Sport, Canberra, ACT, Australia}
\makeatletter
\begin{abstract}
  We introduce a novel deep learning based group activity recognition approach called the Pose Only Group Activity Recognition System (POGARS), designed to use only tracked poses of people to predict the performed group activity.
  In contrast to existing approaches for group activity recognition, POGARS uses 1D CNNs to learn spatiotemporal dynamics of individuals involved in a group activity and forgo learning features from pixel data. 
  The proposed model uses a spatial and temporal attention mechanism to infer person-wise importance and multi-task learning for simultaneously performing group and individual action classification.  
  Experimental results confirm that POGARS achieves highly competitive results compared to state-of-the-art methods on a widely used public volleyball dataset despite only using tracked pose as input.
  Further our experiments show by using pose only as input, POGARS has better generalization capabilities compared to methods that use RGB as input.
\end{abstract}
\maketitle

\section{Introduction}
Developing methods for recognizing activities performed by groups of people is an important computer vision research area applicable for many domains, with two of the most prominent tasks being analysis of sports videos and monitoring of surveillance video footage. 
Examples in the sports domain  include identifying events such as shots on goal in soccer matches, and pick and roll events in basketball. 
Examples in surveillance videos include people fighting, walking together, people being followed and other similar human activities. 
The major difference between group activity recognition and individual action classification is the need to simultaneously reason about multiple people. 
Thus, group activity recognition methods should be specifically designed to understand spatial and temporal evolution of pose for groups of people. 

It is common for existing work on group activity recognition domain to operate on video pixel data directly, 
but thus far it has remained an open question as to how well group activity can be predicted from extracted human poses. 
The objective of our research is to explore the predictive power of pose data by developing a deep learning model that predicts a volleyball group activity label only from tracked pose, without directly using RGB pixel data.

Majority of the existing group activity recognition models utilize person appearance features extracted via Convolutional Neural Network (CNN) based models to learn their spatiotemporal dynamics throughout consecutive frames in the video instance \cite{Ibrahim2016_hierarchical_deep,Ramanathan2016_keyperson,Lu2019_spatioTempAtt}. 
Variants of Recurrent Neural Networks (RNNs) such as Long Short Term Memory (LSTM) networks or Gated Recurrent Units (GRUs) are most widely used to model temporal relationships of activities. 
To better map the relationships between individuals in the scene, some such existing works have introduced pooling strategies and other interpersonal processing techniques \cite{Bagautdinov2017_socialScene,Azar2018_multistream}. 

In contrast to most existing methods which are focused on taking RGB features as input \cite{Ibrahim2016_hierarchical_deep,Ramanathan2016_keyperson} or RGB with additions like optical flow \cite{Azar2018_multistream,Azar2019_convoRelational} and/or pose \cite{Lu2019_spatioTempAtt,Gavrilyuk2020_ActorTransformer}, proposed method uses only pose to predict the group activity label.

Extracting useful high level information such as pose and tracking data from video is a common practice in sports domain \cite{Baradel2017_poseConditioned,Carreira2017_kinetics,Mehrasa2018_DeepTrajectory,Zhu2018_actionMachine}. 
Using pose to represent human subjects has several advantages. 
Pose representations ignore less relevant factors in the input data such as person appearance features and background which allows the models trained using pose to generalize better to new situations like sporting events involving previously unseen teams and venues. 
Ignoring person level identifiable appearance features allows pose data to preserve privacy of the people whose data is being captured. 
In addition, recent commercial computer vision systems such as Microsoft Azure Kinect are capable of directly producing pose information as a better representation of human dynamics in video streams. 
As more researchers realise the usefulness of pose information and methods for capturing such data become more readily available, we anticipate that pose information will be collected for a range of use cases such as analysing basketball player movements, motion tracking and other similar human activities. 
Our method can then take advantage of the already available pose information to more efficiently perform group action recognition. 
This is due to the much lower dimensionality of pose information (usually just 16 key points per person) compared to RGB images.

In addition to pose information, Perez et al.\cite{Perez2020_SkeletonBased_GIRN} recently explored the usage of related object (i.e. ball) trajectories for identifying group activities in sports videos, finding that the additional input information can substantially improve classification accuracy. 
Our experiments partially support this finding, with the inclusion of ball tracklets exhibiting slightly improved empirical results. 
However, our approach is not reliant on ball trajectories to achieve high classification accuracy.

In this study we propose Pose Only Group Activity Recognition System (POGARS), a novel system for performing group activity recognition by utilizing extracted tracked pose as input. 
POGARS uses 1D CNNs to learn the spatiotemporal dynamics of individuals using their pose keypoint estimations and position tracklets. 
We found 1D CNNs were able to model the temporal dynamics in the video more effectively than the RNNs used by most existing methods for group activity recognition.

In most group activity settings, there are certain people who play more central roles than others, and are hence more predictive of the group activity recognition task. 
For example, when recognizing a spike in volleyball we expect more attention should be placed on the players directly involved with spiking and blocking the ball. 
POGARS uses a spatial self attention mechanism for identifying the importance of each individual for the particular group activity. In addition, temporal attention is also important. 
For instance, video frames at which the volleyball is passed can be considered as an important temporal instance in the video clip. 
Proposed approach uses a temporal self attention mechanism to assign different importance weights to the different video frames.

While recognizing group activity of the video instance, we simultaneously perform individual action recognition as an additional task, thus making our model perform multitask learning. 
This was found to result in a noticeable accuracy improvement with respect to the primary task of group activity recognition.

Though group activity recognition is an important research area in computer vision, there are  few video datasets available for model evaluation. 
The volleyball dataset introduced by Ibrahim et al.\cite{Ibrahim2015_volleyballPaper} is one of the most popular  group activity recognition datasets. 
Existing works such as \cite{Ibrahim2016_hierarchical_deep,Azar2018_multistream,Azar2019_convoRelational,Lu2019_spatioTempAtt} have used the volleyball dataset to evaluate group activity recognition models. 
In order to evaluate and compare our proposed approach with the past related work, we have conducted our experiments on  the volleyball dataset.

POGARS achieves 93.2\% accuracy for group activity recognition by only using tracked pose information on the volleyball dataset. 
To contextualize this result, the state-of-the-art actor-transformer model proposed by Guvrilyuk et al.\cite{Gavrilyuk2020_ActorTransformer} achieves 92.3\% when using pose only (their state-of-the-art accuracy of 94.4\% requires optical flow data as an additional input modality).
Another very important benefit of  only using tracked pose is that the trained model can generalize better to testing data with different characteristics by ignoring such factors as lighting conditions, team uniform colour, the colour of the court, audience appearance, etc.
To test this hypothesis we created a skewed train/test split of the volleyball data where the training data consisted of only games played in a single venue in the London 2012 Olympics and the test set consisted of matches from all other venues. 
The results show POGARS's accuracy only dropped slightly from 93.2\% to 89.7\% when using the skewed data. 
In contrast, a competitive I3D model \cite{Carreira2017_kinetics} trained using RGB images as input showed a very large drop in accuracy, from 84.6\% to 73.9\%.

Our contributions can be summarized as follows:
\begin{itemize}
  \item We propose POGARS, the first group activity recognition system that solely utilizes tracked poses as input. We showed a POGARS model trained on a single venue generalizes significantly better to different test venues compared to a competitive model trained on RGB.
  \item We performed a thorough architecture search and found a 1D CNN based approach that uses spatial and temporal attention outperformed the traditional approach of using RNNs for group activity recognition.
  \item We conducted experiments on a popular group activity recognition dataset to show that POGARS achieves near state-of-the- art accuracy by only using tracked pose for model input.
\end{itemize}

\section{Related work}
\label{sec:Related-work}
In this section we review existing work in the area of group activity recognition and the use of deep learning techniques to solve the problem.

Group activity recognition considers the multi-person behavioral and interaction dynamics to interpret the collective activity instances. 
Most of the early work on group activity recognition is based on models that learn information in video frames using hand-crafted features. 
Hierarchical graphical models \cite{Lan2012_discriminativeLatentModels,Amer2014_hirf} and dynamic Bayesian networks \cite{Zhu2013_contextAware} are some of the popular methods used for interpreting group activity in video. 
Choi et al. \cite{Choi2009_colllectiveActivity} introduced a spatio-temporal feature descriptor  based on shape context \cite{Belongie2002_shapeMatching} for interpreting collective activities of people. 
Hierarchical models such as \cite{Lan2012_discriminativeLatentModels} interpreted the group activity by representing the actions of each individual person as well as person to person interactions using a tree-like graphical descriptor. 
Handcrafted feature engineering approaches for collective activity recognition come with various limitations such as domain dependency, large computational costs and poorer resultant accuracy.

The focus of recent collective activity recognition models has shifted towards using deep neural networks \cite{Wu2017_hac_review}. 
Deng et al. \cite{Deng_2016_structureInference} introduced Structure Inference Machine , which has a RNN based backbone for analyzing interactions between people in the scene in order to recognize group activity. 
Ibrahim et al. \cite{Ibrahim2016_hierarchical_deep} proposed a group activity recognition model that consists of two LSTM based sub models where the first one encodes the individual player level actions and their temporal dynamics while a second model fuses outputs from the first model for providing temporal dynamics of the group activity. 
Inspired by this work, two-level RNN based hierarchical methods have become the most popular approach in the domain of collective activity recognition \cite{Ibrahim2016_hierarchical_deep,Ramanathan2016_keyperson,Tsunoda2017_footballAction,Qi2019_stagNet}. 
Though these models learn the spatial features of each person in the scene, most of them don't consider the spatial position of the individuals. 
In contrast to the above existing work, our proposed model uses person level pose and spatial location information (position tracklets) for predicting the collective activity since combination of pose and location information is a dense information representation.

In order to learn relations between individuals, Wu et al.\cite{Wu2019_ARG} employed graph convolutional networks in their Actor Relation Graph (ARG) designed for identifying group activities. 
2D CNNs are used to extract features from person bounding boxes which are used to capture the appearance and position relations between actors by ARG. 
In contrast to ARG, our proposed model is able to learn the temporal relationship of the individuals in the group activity through the inclusion of 1D convolutional layers.

Azar et al. \cite{Azar2018_multistream} proposed a multi-stream CNN framework  for group activity recognition which uses optical flow, warped optical flow, pose heatmaps and the RGB frame as different input modalities for the predictive model. 
Features extracted from CNNs are first applied to each input stream and then fused to produce the group activity prediction for video instances. 
Azar et al. \cite{Azar2019_convoRelational} recently introduced Convolutional Relational Machine (CRM). 
Emphasizing the importance of spatial position information of individuals in group activity classification, the authors have proposed to learn the spatial relations between the activities of people using activity maps in CRM. 
Similarly to \cite{Azar2018_multistream}, CRM also uses optical flow information to enhance prediction accuracy.

Following a similar approach of using CNNs to predict group activities in videos, Gavrilyuk et al.\cite{Gavrilyuk2020_ActorTransformer} introduced an actor-transformer model which uses optical flow for representing temporal dynamics while pose information has been used for interpreting spatial information of the people. 
Our method does not use optical flow information as input due to the high computational costs.

\subsection{Pose-based action recognition}
\label{ssec:pose-based-action-recognition}

Pose data is lightweight and more robust to visual clutter and  noise compared to RGB data. 
Hence, many individual action recognition models use different forms of pose data representations as input feature embeddings \cite{Liu2016_SpatioTemporalLW,Baradel2017_poseConditioned,Wang2017_modelingTemDynamics,Zhu2018_actionMachine}. 
Zhu et al. \cite{Zhu2018_actionMachine} introduced ``Action Machine" which first feeds cropped images of people as input into an I3D model \cite{Carreira2017_kinetics}. 
The model has a RGB stream and also a pose stream. 
These two streams are fused together to predict action classes. 
In contrast to these works (which perform individual action recognition), we perform group activity recognition using tracked pose data.

Inspired by the success of pose-based individual action recognition, Lu et al. \cite{Lu2019_spatioTempAtt} proposed a group activity recognition model built from pose skeleton data and deep RGB features of individuals in each video frame. 
Temporal dynamics of group activity were learnt using GRUs.

Perez et al.\cite{Perez2020_SkeletonBased_GIRN} recently proposed a skeleton-based relational reasoning model for group activity analysis which uses pose skeleton data and ball tracklets for learning interactions between individuals in related objects.

When compared to all the models in this section we differ in  the following ways: 1) we use only pose data instead of both  RGB and pose data; and 2) we are the only solution that uses  1D CNNs to model temporal patterns.

\subsection{Attention models}
\label{ssec:attention-models}

Attention has been successfully applied to both computer vision and natural language understanding domains \cite{Mnih2014_RNNforAtt,Xu2015_showAttendTell}. 
In the most recent studies, collective activity recognition models have used attention mechanisms either as spatial attention or temporal attention. 
Ramanathan et al. \cite{Ramanathan2016_keyperson} introduced the concept of key actors to understand collective activity. 
A bidirectional LSTM based model was used to represent the tracks of people in the scene and the model is capable of attending for a subset of the people actively engaged in collective activity. 
The latent embedding model introduced by Tang et al. \cite{Tang2017_latentEmbedding} consisted of an attention mechanism to encode the importance of each individual person in the collective activity scenario.

In order to identify video frames which make more contribution towards the prediction, different temporal attention approaches have been used in group activity recognition problem domain \cite{varona2000_automaticKeyFrame,Raptis2013_poselet_keyFrame}.
Attention mechanisms employed in these models were designed to extract a small subset of predictive frames from the whole temporal span of the video. 
Instead of making a hard decision to include frames as these works do (so-called ``hard attention"), we use a soft temporal attention mechanism in POGARS which weights the contribution of features from each temporal instant.

Lu et al.\cite{Lu2019_spatioTempAtt} introduced an attention based collective activity recognition model which uses pose and RGB features of each person in the scene to predict the group activity label. 
It combines both spatial and temporal attention mechanisms. 
Whilst RGB data was the primary input for their model, the authors used pose features for determining the distribution of attention among the people.
Our work differs from \cite{Lu2019_spatioTempAtt} in the following ways: 1) we use 1D CNNs instead of RNNs to extract temporal patterns; and 2) we only use pose as input instead of both pose and RGB as attention inputs.

\section{Pose only group activity recognition system (POGARS)}
\label{section-pogars}

\begin{figure}[pos=t]	
	\centering
	\includegraphics[width=1.0\linewidth]{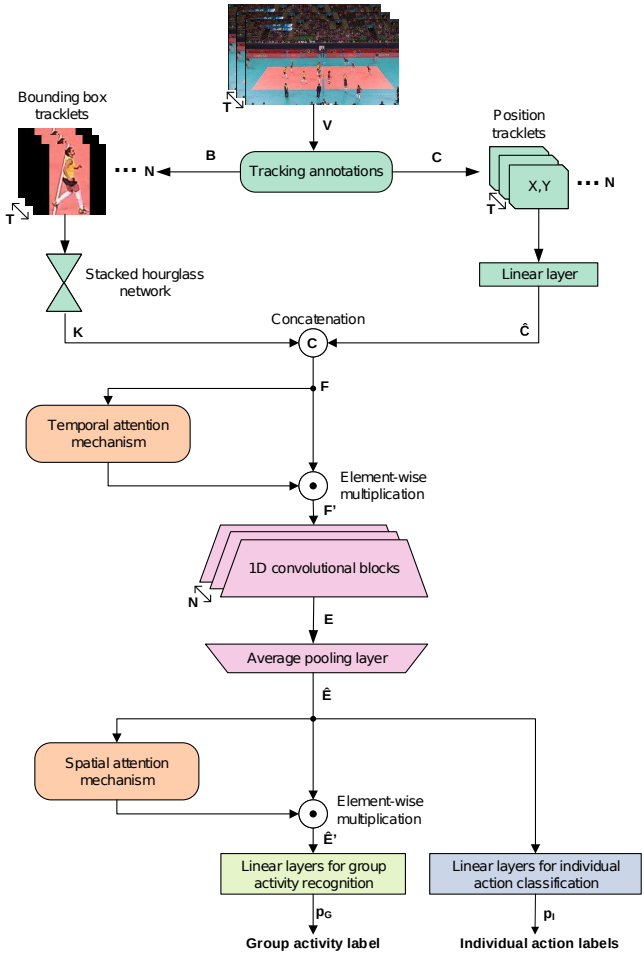}
	\caption{Overview of pose only group activity recognition system (POGARS).}
	\label{fig:pogars-model}
\end{figure}

In comparison with raw image frames, tracked poses of the individuals from the video contain a much denser set of important information allowing the deep learning model to focus only on the most important features. 
Such as a person who is about to perform a spike in volleyball has a vastly different pose from someone who is running for the ball. 
Pose keypoints can be represented as numerical coordinates, as opposed to the high-dimensional pixel data of an RGB image.

One of the biggest challenges for modeling data in video is finding temporal coherence (tracking where one point in one frame moves to the next frame). Modeling temporal coherence in sequences of poses is more natural than in video data since each keypoint is assigned to a different dimension in the pose vector. 
Hence 1D CNNs can effectively model the changes in position of each keypoint across time. 
Further highlighting the most important data for activity recognition, we also use a self attention mechanism to assign a higher weight to more important people and frames for accurate group activity classification. Using extracted pose features instead of RGB features can produce a model that is better able to generalize across datasets since pose excludes factors from the input, such as lighting conditions, team uniform colour, the colour of the court, audience appearance and other nuisance factors.
  
Figure \ref{fig:pogars-model} shows an overview of the proposed POGARS approach. 
First, position coordinates and associated bounding boxes of each person across time in the input video are acquired from manual annotations used in \cite{Sendo2019_HeatmappingOfPeople}.
Next we generate 16 2D keypoint estimations for each individual by feeding the bounding box tracklets to stacked hourglass human pose estimation algorithm \cite{Newell2016_stackedHourGlass}. 
Position coordinate tracklets are fed into a fully connected layer to create a descriptive feature embedding for the position of each individual in the frame. 
Position embeddings and keypoint pose representations are concatenated together to produce a composite feature embedding for each individual that contains pose and position information.

Feature representations of consecutive frames in the video clip contribute in different degrees to class prediction. 
The concatenated feature embeddings of each person in each frame is fed into a temporal attention mechanism to calculate frame-wise attention weights. 
An element-wise multiplication between the feature embedding and attention weights across the temporal dimension is performed before feeding the features into a set of 1D CNN blocks.
Subsequent CNN blocks with residual connections are able to learn temporal dynamics of the individuals using the pose and position embeddings.

POGARS performs multi-task learning by predicting both group activity label $p_G$ and individual action labels $p_I$. 
This is achieved by attaching two separate heads (one for group activity prediction and the other for individual activity prediction) to the output feature set from 1D CNN blocks.

To implement spatial attention, attention weights of each individual are multiplied element-wise with the corresponding feature embedding to ensure the feature embeddings of key people in the video get more attention in the process of predicting the group activity. 
Detailed descriptions of each component of our method are presented in the following subsections.

\subsection{Tracking and pose estimation}
\label{ssec:tracking-pose-estimation}

Since our collective activity recognition model is based on position and pose feature representations of individuals, we employ manually annotated bounding boxes of the people from \cite{Sendo2019_HeatmappingOfPeople} which are in the image space. 
Our approach decouples tracking and pose estimation from the model used to classify the group activities. 
In the event that manually annotated tracks are not available any multi-object detection and tracking algorithm can be used, with the final classification accuracy expected to correlate positively with the performance of these algorithms.

After object detection, we extract the human pose of each person. 
There are many existing 2D human pose estimation models which are appropriate for inferring such pose keypoints \cite{Shotton2011_realtimePose,Cao2016_multiPersonPose,Newell2016_stackedHourGlass}. 
Pose estimation methods which first detect body joints and then group them to form individual poses per person are called bottom-up approaches. 
In contrast, top-down approaches first detect people in the scene and then predict keypoints of each detected person. 
Top-down approaches have been found to be more accurate than bottom up approaches for performing pose estimation \cite{He2017_mask-rcnn}. 
Hence, we take a top-down approach by first using \cite{Danelljan2014_robustVisTracking} to find bounding boxes around the people and then employing the pre-trained stacked hourglass model \cite{Newell2016_stackedHourGlass} to estimate joint coordinates of each person. 
Due to the modular nature of our group activity recognition method, any human tracking algorithm and 2D human pose estimation algorithm can be used with our proposed approach. 
This includes bottom-up pose tracking approaches, which may be suitable candidates for situations where computational speed is of key importance.

Given an input video clip \textit{$V$} containing \textit{$N$} people and spanning \textit{$T$} frames,
manual annotations have been used to generate \textit{$N$} bounding box tracklets \textit{$B\in \mathbb{R}^{N\times 4\times T}$} and position coordinate tracklets of each bounding box \textit{$C \in \mathbb{R}^{N\times 2\times T}$}.

For each individual, 16 2D pose keypoints \textit{$K\in \mathbb{R}^{N\times 32\times T}$} are estimated using the stacked hourglass pose estimation algorithm.
A linear layer is added on top of the bounding box position coordinates $C$ in order to generate a descriptive feature embedding \textit{$\hat{C}\in \mathbb{R}^{N\times 32\times T}$} to represent the spatial position of the people in each frame.
Pose feature representation $K$ and position feature representation $\hat{C}$ of each corresponding individual are concatenated, so \textit{$(K|\hat{C}) = F\in \mathbb{R}^{N\times 64\times T}$} feature embedding of the tracked pose is generated.
Since the temporal evolution block is responsible for learning each person level feature separately, the order of people is important. 
Thus we ensured that the pose feature representations of each individual in each frame are ordered consistently by stacking them in a left to right in terms of x-position in the video frame.

\subsection{POGARS with different person level fusion and temporal modeling  approaches}
\label{ssec:fusion-approaches}

\begin{figure}[pos=t]
  \centering
  \includegraphics[width=1.05\linewidth]{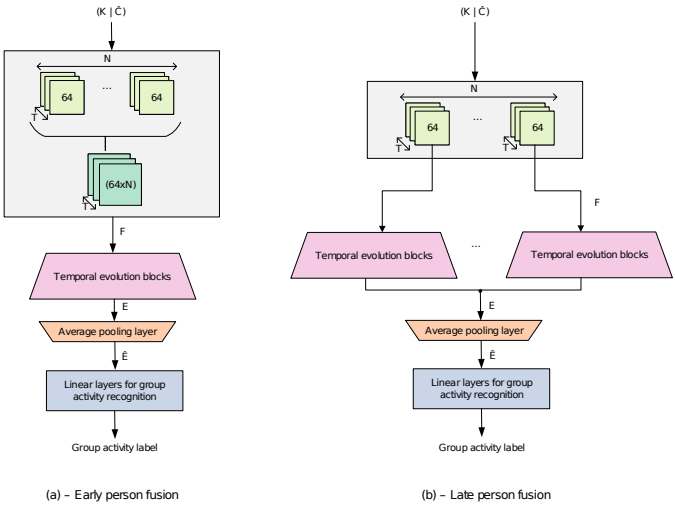}
  \caption{Different person level fusion approaches of POGARS. Each approach was tested with two temporal evolution modeling variations: bidirectional LSTMs and 1D CNNs.}
  \label{fig:fusion-approaches}
\end{figure}

POGARS captures the spatial and temporal evolution of  individuals in order to predict the group activity label. 
Person level spatiotemporal features are fused together to generate a person level feature representation. 
Learning group activities in the video instances is done by modeling temporal evolution of the fused feature representations.  
We explored two different person level fusion strategies and two temporal modeling variations within POGARS.

\subsubsection{Person-level fusion}
\label{sssec:person-level-fusion}

Figure \ref{fig:fusion-approaches} illustrates the two different person level fusion approaches we explored. 
These two approaches are early person fusion and late person fusion. 
In \textbf{early person fusion}, person-level feature independence is removed before modeling the temporal evolution by fusing the tracked pose features of all people before input into the module that captures temporal dynamics. 
Alternatively, in \textbf{late person fusion} the temporal evolution of each person is first modeled individually before the learned features are fused together at the end.

In the late person fusion method, the temporal evolution block is responsible for learning the tracked pose of each person separately while the early person fusion method combines person level feature embeddings of each frame before combining the temporal dimension using the temporal modeling block. 
The early person fusion approach eliminates the need for person level tracking by combining the features of each person before temporal modeling. 
Although not needing to perform tracking eliminates one potential source of error, our experiments show person-wise spatiotemporal feature learning (as provided by late person fusion) is essential for achieving higher classification accuracy.

\subsubsection{Temporal evolution modeling}
\label{sssec:temporal-evolution-modeling}

Most of the existing group activity recognition mechanisms \cite{Deng_2016_structureInference,Ibrahim2016_hierarchical_deep,Tsunoda2017_footballAction,Lu2019_spatioTempAtt} use LSTM or GRU based networks for learning temporal dynamics of the video. 
In our experiments we evaluated two temporal evolution modeling  approaches: bidirectional LSTMs and 1D CNN, and found that POGARS networks which utilise 1D convolutions outperform their LSTM counterparts. 
We give a detailed description of our 1D CNN based temporal modeling block in section \ref{ssec:temporal-evolutions-individuals}.

\subsection{Temporal attention mechanism}
\label{ssec:temporal-attention-mechanism}

Group activities have different important stages in their temporal span. 
For example, in a volleyball spiking activity, the frames in close temporal proximity to the instant that the player strikes the ball are typically more important than rest of the frames. 
Directing the model to attend to features from the most important frames prevents the model from overfitting to less important frames which may contain irrelevant information. 
Hard attention based methods such as \cite{varona2000_automaticKeyFrame,Yan2018_deepKeyFrame} detect a specific set of video frames that maximally contribute to the final prediction. 
Rather than selecting a small number of frames to keep for further analysis and discarding the rest, we propose a soft self attention mechanism which assigns an importance weight to every frame.

The attention mechanism we have employed in POGARS contains a set of linear layers and a softmax layer.
Given the feature embedding of individuals $F$ , temporal attention is computed as,

\begin{equation}
  F' = softmax(\phi (F)) \odot F
\end{equation}

where $\phi$ is the temporal attention function (modelled using linear layers).
Computed temporal attention weights are multiplied element wise across the time dimension with the feature embedding $F$.
This produces \textit{$F'\in \mathbb{R}^{N\times 64\times T}$}which is the feature embedding of the video clip weighted according to frame-wise temporal importance.

\subsection{Temporal evolutions of individuals based on tracked pose}
\label{ssec:temporal-evolutions-individuals}

\begin{figure}[pos=t]
  \centering
  \includegraphics[width=1.0\linewidth]{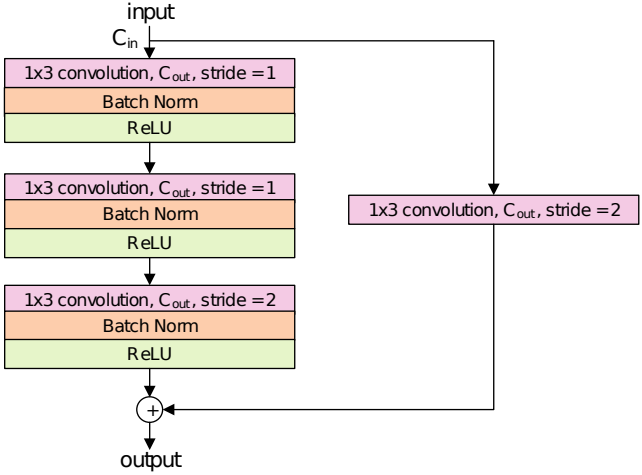}
  \caption{Composition of a single convolutional block of POGARS.}
  \label{fig:convolutional-block}
\end{figure}

In the proposed method, people in the video clip are represented by feature embeddings of their respective tracked poses. 
Learning temporal evolution of the pose and position dynamics of individuals is critical for predicting the collective activity performed by the group. 
We employed 1D CNNs to model the temporal evolutions of individuals instead of RNNs, based on the study we did with different POGARS architectures. 
In contrast to RNNs, the ability of 1D CNNs to learn deep related temporal features and translational equivariance makes them a better fit for representing fine grained pose and position evolutions of individuals across time. 
Figure \ref{fig:convolutional-block} shows the composition of convolutional blocks used in POGARS.
Each convolutional block consists of 3 1D CNN layers with a skip connection, and POGARS contains 4 such blocks stacked together.
The stack takes the post-attention feature embeddings $F'$ as input and analyses the motion of individuals by producing a temporally convolved feature representation {$E\in \mathbb{R}^{N\times 1024\times T}$}.

\subsection{Spatial attention mechanism}
\label{ssec:spatial-attention-mechanism}

In POGARS we employ a spatial attention mechanism that uses temporally convolved feature extractions of each individual to identify each person's importance score for the particular group activity.
Our spatial attention block contains a set of linear layers and a softmax layer.
Temporally convolved feature embeddings $E$ are average pooled in the time dimension to produce a feature embedding for each person in the video clip {$\hat{E}\in \mathbb{R}^{N\times 1024\times 1}$} where $N$ is the number of individuals in the video.
The spatial attention is computed as,

\begin{equation}
  \hat{E'} = softmax(\psi (\hat{E})) \odot \hat{E}
\end{equation}

where $\psi$ is the spatial attention function.
Computed spatial attention weights are multiplied element wise across the person dimension with the feature embedding $\hat{E}$.
This produces {$\hat{E'}\in \mathbb{R}^{N\times 1024\times 1}$} which is the feature embedding of all individuals in the video clip weighted accordingly to each person's importance.

\subsection{Multi-task learning for group activity recognition and individual action recognition}
\label{ssec:multi-task-learning}

In parallel to learning the collective activity label, POGARS uses parameter sharing to also predict the individual actions of each person in the video clip.
The average pooled feature representation is shared between two task-specific linear layer stacks.
One of these layers predicts the group activity label, whilst the other predicts individual action labels for each person.
Feature embeddings used for individual activity recognition are not weighted with the spatial attention mechanism.

In order to train our proposed multi-task model, the losses of group activity recognition task and individual action recognition are optimized simultaneously.
Considering $\mathcal{L}_{IA}$ as the individual action prediction loss and $\mathcal{L}_{GA}$ as the group activity recognition loss, the multi-task loss function $\mathcal{L}_{MT}$ of the model is defined as follows:

\begin{equation}
  \mathcal{L}_{MT} = \alpha \mathcal{L}_{IA} + \mathcal{L}_{GA}
  \label{eq-combined-loss}
\end{equation}

We used multi-class cross entropy loss as the loss function for both individual and group activity prediction in our experiments.
To achieve the optimum accuracy for group activity recognition, we found that $\alpha$ set to 2.0 gave the best results.

\subsection{ Relationship between individuals and relevant objects for group activity recognition}
\label{ssec:ball-indv-relationship}

\begin{figure}
  \includegraphics[width=0.45\textwidth]{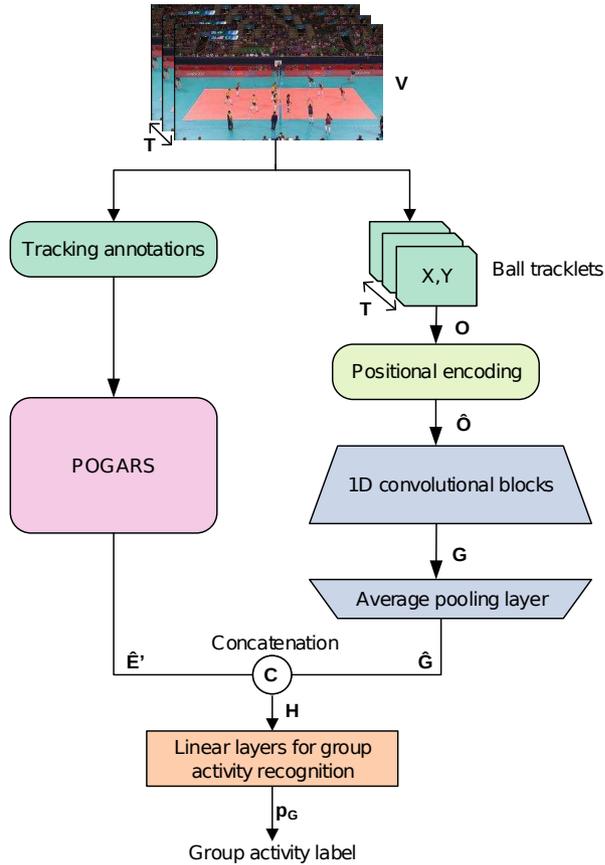}
  \caption{Adding ball as an input modality for POGARS.}
  \label{fig:pogars-ball}
\end{figure}

Using specific activity-related objects such as a ball, bat or a puck is a common scenario in group sports. 
Perez et al.\cite{Perez2020_SkeletonBased_GIRN} explored the effectiveness of learning ball tracklets for predicting group activity instances in volleyball videos using their GIRN model. 
In order to benchmark POGARS with GIRN, we experimented with adding position coordinates of the volleyball as an input modality for our network (See figure \ref{fig:pogars-ball}).

Given the ball position coordinates in the image space \textit{$O\in \mathbb{R}^{2\times T}$} , sinusoidal positional encoding \cite{Vaswani2017_attentionAll} is applied in order to generate a descriptive feature embedding \textit{$\hat{O}\in \mathbb{R}^{64\times T}$}. 
The feature embeddings are fed into a stack of 1D CNN blocks (similar to the architecture described in section \ref{ssec:temporal-evolutions-individuals}) to produce a temporally convolved feature representation \textit{$G\in \mathbb{R}^{1024\times T}$}, followed by global average pooling to remove the temporal dimension. 
The resulting ball features are concatenated to $\hat{E'}$ for generating \textit{$H\in \mathbb{R}^{(N+1)\times 1024}$}, which is employed for predicting group activities by using a linear layer stack.

\section{Experiments}
\label{sec:experiments}

\subsection{Experimental setup}
\label{ssec:experimental-setup}

POGARS is evaluated by performing experiments on the popular volleyball dataset \cite{Ibrahim2015_volleyballPaper}. 
Manually annotated bounding box tracklets and position coordinate tracklets of each player in the volleyball videos are obtained from \cite{Sendo2019_HeatmappingOfPeople}. 
The annotations are in image space.
We utilized stacked hourglass pose keypoint estimation algorithm \cite{Newell2016_stackedHourGlass} for predicting 16 keypoints \textit{(ankles, knees, hips, pelvis, spine, neck, head, wrists, elbows and shoulders)} of each player in the video frames. 
The stacked hourglass network has been pretrained using the MPII dataset \cite{Andriluka2014_humanpose}. 
Coordinates of the extracted keypoints are expressed relative to the pelvis joint location, whereas the position coordinate tracklets are expressed in absolute frame space.

Coordinate tracklets of the people are scaled to values between 0-1.
From each video clip, 36 frames are utilized in model training and evaluation.
We used stochastic gradient descent with ADAM \cite{Kingma2014_adam} with initial learning rate set to $10^{-3}$, step-wise learning rate decay with a step size of 7 and fixed moving average decay rates $\beta_{1} = 0.9, \beta_{2} = 0.999$.
Models are implemented using the PyTorch deep learning framework \cite{Paszke2019_PyTorch} and trained on an NVIDIA GeForce RTX 2080 Ti GPU.

\subsection{Volleyball dataset}
\label{subsection-volleyball-dataset}

The volleyball dataset introduced by Ibrahim et al. \cite{Ibrahim2015_volleyballPaper} contains 55 videos collected from publicly available YouTube volleyball matches. 
It includes 4830 trimmed group activity instances which belong to 8 activity classes. 
Each activity instance contains 41 frames where the middle frame is labeled with the group activity label and individual action labels. 
The 9 individual action labels are \textit{spiking, blocking, setting, jumping, digging, standing, falling, waiting,} and \textit{moving}. 
The 8 group activity labels are \textit{right spike, left spike, right set, left set, right pass, left pass, right winpoint,} and \textit{left winpoint}.

For our experiments we use the same manual bounding box annotations and player positions as \cite{Sendo2019_HeatmappingOfPeople}.
We have also followed the train/test splits suggested by \cite{Ibrahim2015_volleyballPaper}. 
In order to reduce model overfitting, we perform data augmentation by also training on horizontally flipped versions of examples from the training set and flipping the activity labels accordingly.

\subsection{Experiment results}
\label{ssec:experiment-results}

\subsubsection{Ablation study}
\label{sssec:ablation-study}

\begin{table}[pos=t]
  \caption{Comparison of different settings of the POGARS evaluated on Volleyball dataset.}
  \label{tab:ablation-study}
  \resizebox{.5\textwidth}{!}{%
    \begin{tabular}{lc}
      \hline
      \textbf{Method}                                                               & \textbf{Accuracy} \\ \hline\noalign{\smallskip}
      POGARS - early person fusion setting with LSTM based network                  & 81.4              \\
      POGARS - late person fusion setting with LSTM based network                   & 85.9              \\
      POGARS - early person fusion setting with 1D CNN based network                & 83.0              \\
      POGARS - late person fusion setting with 1D CNN based network                 & 88.3              \\  \hdashline\noalign{\smallskip}
      POGARS - with only spatial attention                                          & 90.8              \\
      POGARS - with only temporal attention                                         & 91.7              \\
      POGARS - with spatial and temporal attention                                  & 92.1              \\ \hdashline\noalign{\smallskip}
      \textbf{POGARS ALL - with spatial, temporal attention and multitask learning} & \textbf{93.2}     \\ \hline
    \end{tabular}%
  }
\end{table}

POGARS consists of two different attention mechanisms and a fusion mechanism that promotes learning spatiotemporal dynamics of the individuals involved with the activity. 
We performed an ablation study to assess the impact of these key features on overall performance. 
Moreover we performed experiments on two different temporal modeling methods for POGARS.

As the results in table \ref{tab:ablation-study} show, both spatial and temporal attention are beneficial and complementary. 
Furthermore, the results show that the late person fusion approach achieves higher accuracy than early fusion. 
Consequently we use late person fusion for the remainder of our experiments.

Instead of RNNs, we used 1D CNNs for the backbone of our model. 
Results show that POGARS using 1D Convolutions (late person fusion without attention) outperforms an LSTM-based counterpart model with the same basic architecture by 2.4\%. 
This demonstrates the ability of 1D CNNs to learn lower-level fine grained feature dynamics of individuals. 
Moreover, we theorize that the translational equivariance of CNNs is useful for understanding repetitive movement dynamics of group activities.

While predicting the group activity label, our model is capable of predicting the individual action labels of people in the scene with the aid of a multi-task learning approach. 
We used a combined loss function as stated in equation \ref{eq-combined-loss} for model training. 
Superior accuracy of the multi-task model clearly indicates that multi-task learning leads to the implicit development of shared features, and that these features are more predictive of group activity than features learned through group activity recognition alone. 
Though individual action recognition is not the focus of our work, we achieved 79.5\% accuracy for that particular task using pose information only.

\subsubsection{Generalization capability of POGARS}
\label{ssec:generalization-capability}

\begin{table}[pos=t]
  \caption{Test accuracies for volleyball dataset with different train/test splits.}
  \label{tab:generalization-results}
  \resizebox{.48\textwidth}{!}{%
    \begin{tabular}{lccc}
      \hline
      \textbf{Method}                     & \textbf{Default data split}   & \textbf{Olympic data split} & \textbf{Difference} \\ \hline\noalign{\smallskip}
      I3D \cite{Carreira2017_kinetics}    & 84.6                          & 73.9                        & 10.7                \\ 
      POGARS ALL                          & 93.2                          & 89.7                        & 3.5                 \\ \hline\noalign{\smallskip}
    \end{tabular}%
  }
\end{table}

To test the generalization capability of POGARS, we created an experiment in which we split the volleyball dataset according to the match venues. 
29 videos out of the 55 videos in the dataset, were recorded at the  same 2012 London Olympics venue. 
We designated that particular set of videos as the train set while the rest were used for model testing. 
We named this special data split as the \textit{``Olympic split"}. 

Since most of the group activity recognition approaches \cite{Azar2019_convoRelational,Tran2015_C3D,Lu2019_spatioTempAtt} have used 3D convolution based networks, we used a I3D based network \cite{Carreira2017_kinetics} for the comparison with POGARS. 
The I3D network is a well recognized and strong baseline algorithm for action recognition. 
The I3D network we used was pretrained on the Imagenet \cite{simonyan2014_veryDeep} and Kinetics datasets \cite{Carreira2017_kinetics}. 
Table \ref{tab:generalization-results} summarises the validation accuracies of I3D model and POGARS for the different data splits of volleyball dataset.

As shown in Table \ref{tab:generalization-results}, POGARS exhibits only a 3.5 percentage point decrease in test accuracy when training and evaluating on the skewed Olympics data split, as opposed to the much larger drop of 10.7 percentage points exhibited by the I3D model.
This observation implies training a model only using tracked pose (like for POGARS) is better able to generalize across different playing environments and venues compared to using RGB image input to train a model (like I3D model) for group activity recognition.

\subsubsection{Visualization of the spatial and temporal attention mechanisms}
\label{ssec:visualization-attention-mechanisms}

In order to qualitatively assess the effectiveness of the spatial and temporal attention mechanisms, we visualized both spatial and temporal attention outputs of the best performing POGARS.

\begin{figure}[pos=t]
  \includegraphics[width=1.0\linewidth]{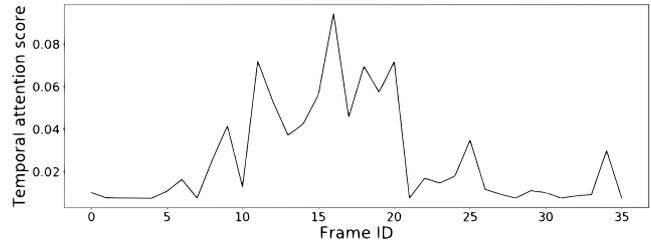}
  \caption{Average attention weights assigned for each temporal instance for the videos in volleyball dataset.}
  \label{fig:temporal-attention}
\end{figure}

\begin{figure*}
  \includegraphics[width=1.0\textwidth]{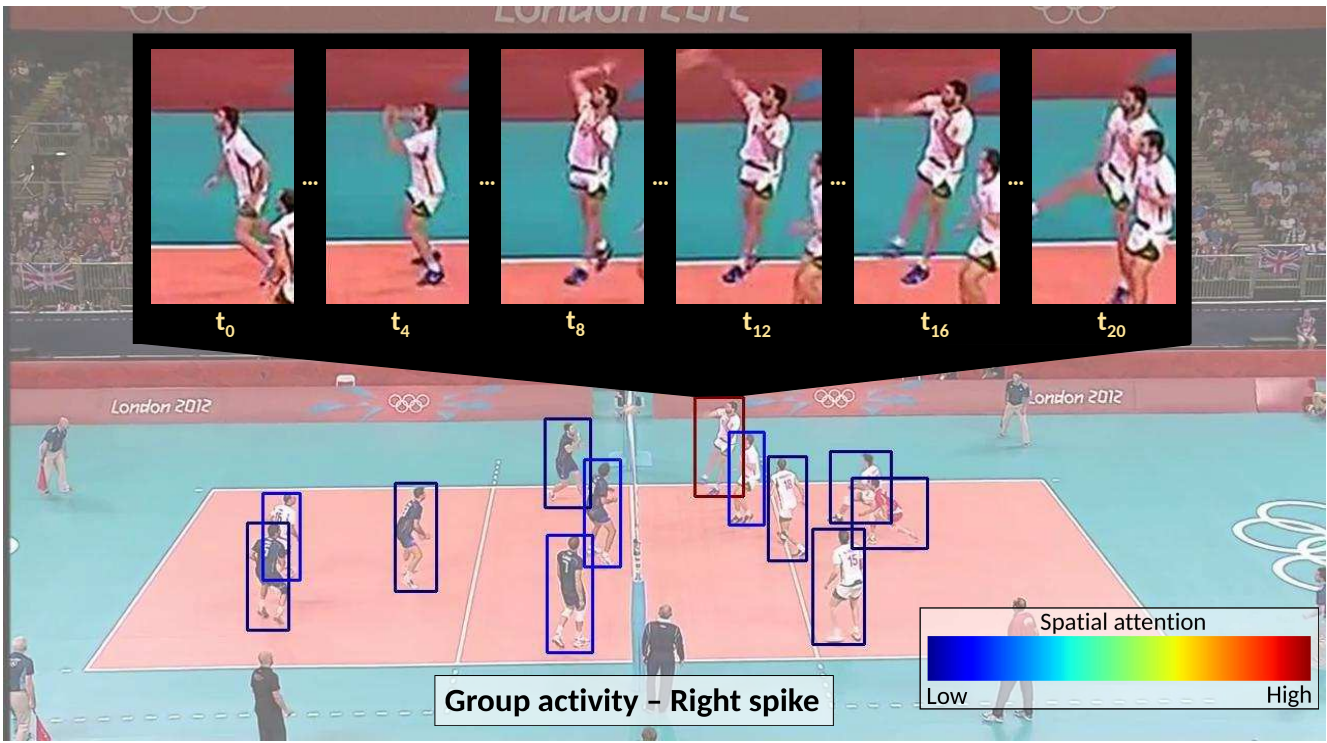}
  \caption{Spatial attention assigned for the players involved in right spike
  activity by POGARS. Key frame of the activity is visualized with player
  bounding boxes colored according to the spatial attention score.}
  \label{fig:spatial-attention}
\end{figure*}

Since group activities have different important stages in their temporal span, our temporal attention mechanism described in section \ref{ssec:temporal-attention-mechanism} is capable of assigning attention weights to each frame in accordance with its perceived importance.
Each activity clip of the volleyball dataset we used in our experiments consists of 36 frames where the 16\textsuperscript{th} frame is the central keyframe of the group activity.
Therefore we expect that frames neighboring the 16\textsuperscript{th} frame ($t=15$) should generally receive high attention when predicting the group activity label.
Figure \ref{fig:temporal-attention} illustrates the predicted mean temporal attention weights for each frame index as assigned by POGARS.
The graph indicates a clear peak in attention weights when $t=15$ (key frame) and comparatively high attention weights for the neighboring frames.
This observation indicates that our temporal attention mechanism can identify the predictive power of frames from group activity video instances.

The spatial self attention mechanism discussed in section \ref{ssec:spatial-attention-mechanism} attends to the key people involved in the group activity. 
Though group activities are labeled according to the side of the volleyball court they were performed from (such as \textit{left spike} and \textit{right spike}), players on both sides of the court contribute to a group activity. 
For example, the players on the left hand side of the court may form a defensive formation when a right spike activity is taking place. 
Figure \ref{fig:spatial-attention} visualizes the attention weights assigned for players in a \textit{right spike} group activity of volleyball data, along with several frame images cropped around the player with highest attention. 
In this case the player with highest attention is the one spiking the ball, so it is clear that POGARS has correctly assigned attention to the major role-player for this particular group activity.

\begin{table}[pos=t]
  \caption{Comparison of different baselines and state-of-the-art methods with POGARS evaluated on volleyball dataset.}
  \label{tab:pogars-comparison}
  \resizebox{.49\textwidth}{!}{%
    \begin{tabular}{lcc}
      \hline
      \textbf{Method}                                                               & \textbf{Feature extraction method} & \textbf{Accuracy} \\ \hline\noalign{\smallskip}
      B1                                                                            & RGB                                & 73.3              \\
      B2                                                                            & RGB                                & 75.1              \\
      B3                                                                            & RGB                                & 63.0              \\ \hdashline\noalign{\smallskip}
      Two-stage hierarchical model \cite{Ibrahim2016_hierarchical_deep}             & RGB                                & 81.9              \\
      CERN \cite{Shu2017_CERN}                                                      & RGB                                & 83.3              \\
      I3D \cite{Carreira2017_kinetics}                                              & RGB                                & 84.6              \\
      Social Scene \cite{Bagautdinov2017_socialScene}                               & RGB                                & 89.9              \\
      Action Relation Graph \cite{Wu2019_ActionRelationGraph}                       & RGB                                & 92.6              \\
      Joint learning with social groups \cite{Ehsanpour2020_JointLearning}          & RGB                                & 93.1              \\ \hdashline\noalign{\smallskip}
      Multi-stream CNN \cite{Azar2018_multistream}                                  & RGB + Optical flow                 & 90.5              \\
      Spatio-temporal attention based model  \cite{Lu2019_spatioTempAtt}            & Pose + RGB                         & 91.7              \\
      Convolutional Relational Machine \cite{Azar2019_convoRelational}              & RGB + Optical flow                 & 93.0              \\
      \textbf{Actor Transformers \cite{Gavrilyuk2020_ActorTransformer}}             & \textbf{Pose + Optical flow}       & \textbf{94.4}     \\ \hdashline\noalign{\smallskip}
      Group Interaction Relational Network \cite{Perez2020_SkeletonBased_GIRN}      & Pose                               & 88.4              \\
      Actor Transformers \cite{Gavrilyuk2020_ActorTransformer}                      & Pose                               & 92.3              \\
      \textit{\textbf{POGARS ALL}}                                                  & \textit{\textbf{Pose}}             & \textit{\textbf{93.2}} \\ \hdashline\noalign{\smallskip}
      Group Interaction Relational Network \cite{Perez2020_SkeletonBased_GIRN}      & Pose + Ball tracklets              & 92.2              \\
  \textit{\textbf{POGARS with ball}}                                                & \textit{\textbf{Pose + Ball tracklets}} & \textit{\textbf{93.9}} \\ \hline
    \end{tabular}%
  }
\end{table}

\subsubsection{Performance comparison of POGARS against rival methods}
\label{sssec:performance-comparison}

We compare the performance of POGARS with state-of-the-art methods and 3 baselines. Table \ref{tab:pogars-comparison} reports the group activity recognition accuracy of the stated models on the volleyball dataset \cite{Ibrahim2015_volleyballPaper}.

The baselines are:

\begin{itemize}
  \item \textbf{B1 - Activity recognition with keyframe spatial features:} This baseline uses the middle frame of the activity video clip (21\textsuperscript{st} frame among 41 frames) as the input to a ImageNet pre-trained ResNet34 model \cite{He2016_Resent}.
  \item \textbf{B2 - Activity recognition with person level spatial and position features:} Similar to B1, except that we introduce a second stream for processing keyframe player bounding boxes represented as heatmaps.
        Learned spatial features from the two streams are fused together late in the architecture to predict the group activity.
  \item \textbf{B3 - Activity recognition with spatiotemporal RGB features:} This baseline takes 16 consecutive frames of the collective activity as input to a network based on ResNet18 architecture with 3D CNN layers.
        B3 can be thought of as a temporally extended version of B1.
\end{itemize}

Both B1 and B2 use keyframe based spatial feature extractions for learning the group activity.
Though these models don't consider the temporal aspect of the activities, they performed reasonably well in collective activity recognition by leveraging state-of-the-art image classification network architectures and transfer learning \cite{Yosinski2014_transferLearning}.
In B2 we use a second input stream consisting of heatmap representations of player bounding boxes in the keyframe.
The slight increase of the accuracy in B2 compared to B1 implies that using player positions as additional input is beneficial in predicting group activity labels.

Compared to B1 and B2, B3 consumes more memory and computational power due to the 3D convolutional layers used in the network. 
The low accuracy of B3 suggests that the inclusion of deep 3D CNN networks is not a favorable trade-off in terms of computation versus accuracy. 
All baselines (B1, B2 and B3) use RGB based feature extraction to learn the spatiotemporal dynamics of the group activities. 
In our proposed group activity recognition model, we utilize pose keypoint estimations of the people and their position coordinates to create a descriptive feature embedding for each individual. 
Compared with the baseline approaches, POGARS reports a significantly higher achieved accuracy as well as faster model convergence.

POGARS performs better than the majority of existing group activity recognition algorithms on the Volleyball dataset (see table \ref{tab:pogars-comparison}). 
This includes methods which use RGB image based feature representations (\cite{Ibrahim2016_hierarchical_deep,Shu2017_CERN}), variants of optical flow (\cite{Azar2018_multistream,Azar2019_convoRelational}), and pose information based networks (\cite{Perez2020_SkeletonBased_GIRN}). 

To our knowledge, the Actor Transformer model \cite{Gavrilyuk2020_ActorTransformer} is the only group activity recognition model that achieves higher accuracy (1.2\% improvement) than POGARS on the volleyball dataset. 
\cite{Gavrilyuk2020_ActorTransformer} makes use of a computationally expensive 3D CNN based architecture which uses optical flow and pose information as input modalities.
Although we recognize that utilizing different input modalities can improve benchmark accuracy results, they require more resource-intensive models to process.

We ran an additional experiment to evaluate the effectiveness of employing tracklets of related objects for group activity recognition by using ball tracklets as an additional input modality for POGARS. 
Our results in table \ref{tab:pogars-comparison} show that POGARS is able to take advantage of the ball tracklets to improve classification accuracy by 0.7 percentage points. 
POGARS with ball tracking outperforms Group Interaction Relational Network\cite{Perez2020_SkeletonBased_GIRN}, the only other existing approach which makes use of ball trajectories.

\begin{figure}[pos=t]
  \centering
  \includegraphics[width=1.0\linewidth]{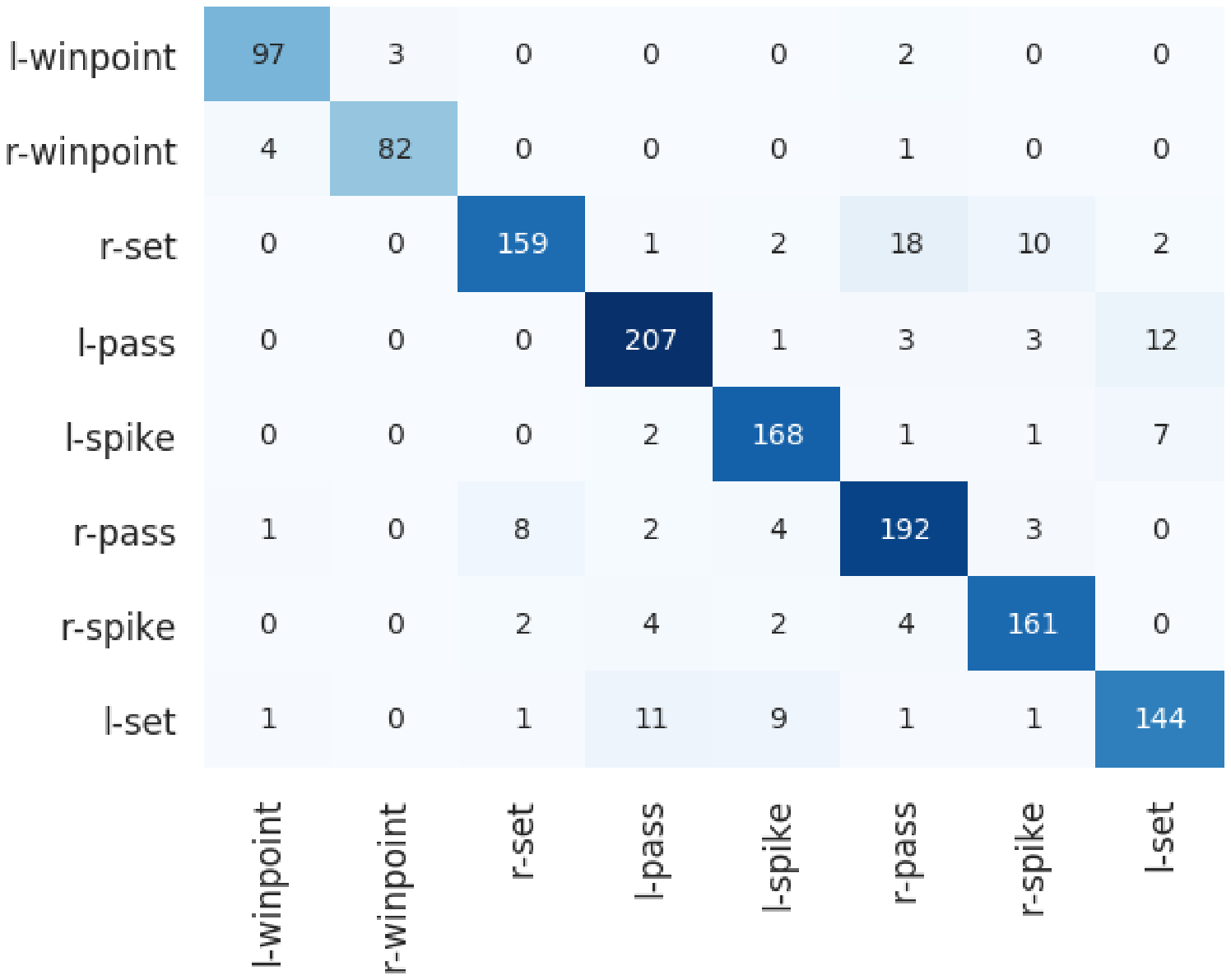}
  \caption{Confusion matrix obtained using best performing POGARS for the volleyball dataset.}
  \label{fig:confusion-matrix}
\end{figure}

Figure \ref{fig:confusion-matrix} contains a confusion matrix summarizing correct and incorrect group activity label predictions.
For the most part, POGARS is capable of providing accurate predictions for all 8 classes in the volleyball dataset. 
It can be seen that the proposed model will occasionally confuse \textit{``pass"} and \textit{``set"} activities, 
which can be explained by the fact that both of these activities are defined by the ball being passed between teammates and hence often involve similar player configurations.

\section{Conclusions}
\label{sec:conclusions}

We propose POGARS, a novel architecture for identifying group activities. 
It does this by learning spatial and temporal dynamics of individuals and fusing learned features to classify the group activity. 
Using only tracked pose as input, POGARS is able to achieve highly competitive results on the Volleyball dataset with a very important benefit of generalizing the model to testing conditions that differ significantly from training conditions. 
We also demonstrated that our method's temporal attention mechanism is capable of identifying the specific importance of each frame in the video clip while the spatial attention mechanism provides a person-wise importance score based on their involvement for the particular collective activity. 
For future work we plan to explore solving the problem of temporal action localization in untrimmed video using only tracked pose. 

\bibliographystyle{cas-model2-names}

\bibliography{cas-refs}

\end{document}